# LoRA-Based Cascaded Multimodal Fusion for Action Recognition in Medical Training Environments

Divya Mereddy[✉*]
divya.mereddy@vanderbilt.edu
Vanderbilt University
Nashville, TN, USA

Jeevan Beedareddy
Quince
Mountainview, CA, USA

## Abstract

This paper presents a cascaded Low-Rank Adaptation (LoRA)-based multimodal fusion framework for action and activity recognition in healthcare-oriented training environments. The proposed architecture combines parameter-efficient modality-specific adaptation with sequential fusion, enabling modalities to be integrated in stages without retraining previously learned components. Rather than assuming a fixed fusion structure, the framework first integrates more closely related modalities and then incorporates additional heterogeneous modalities, supporting scalable adaptation across datasets with different modality sets.We evaluate the framework on two healthcare-oriented training environment datasets: NurViD and the Nurse Training dataset. Across these datasets, preliminary results suggest that the proposed cascaded fusion strategy improves over individual modality models and provides competitive performance relative to previously reported dataset-specific baselines. Overall, these findings indicate that cascaded LoRA-based fusion is a promising parameter-efficient approach for integrating heterogeneous modalities in medical training action and activity recognition tasks. github: https://github.com/anonymous0-ai/LoRA-Based-Cascaded-Multimodal-Fusion-.git.

## 1 Introduction

The ability to accurately recognize and classify human actions is critical across numerous domains, including surveillance, healthcare, education, and training environments. With the increasing complexity of real-world environments, action recognition tasks often require the integration of diverse data modalities, such as skeletal, flow, RGB, and audio information. Multimodal fusion models have emerged as a promising solution to these challenges by leveraging complementary modalities to provide a richer and more nuanced understanding of activities. However, existing approaches face several limitations, including high computational costs, inefficiency in adapting to new modalities, and the need for retraining entire models when datasets evolve.

To address these challenges, we propose a novel **LoRA-Based Cascaded Multi modal Fusion Architecture** that combines the strengths of cascaded fusion with the computational efficiency of LoRA-based fine-tuning. The framework introduces a modular architecture that enables the sequential integration of modalities without retraining previously trained components, offering significant flexibility and scalability. For example, our approach allows for incremental modality addition, such as combining skeleton and RGB data (Stage 1) and subsequently integrating text or other modalities (Stage 2), while preserving the integrity and performance of earlier stages.

The framework is evaluated using two benchmark datasets—**NurViD**, a large-scale video audio dataset focused on action localization in clinical environments, and the **Nurse Training Dataset**, a sensors based domain-specific dataset for healthcare action recognition. By leveraging the complementary strengths of different modalities (such as skeleton, RGB, and textual) in a cascaded fusion setup, our approach demonstrates promising results (specially with limited resources) compared to state-of-the-art fusion methods. Specifically, our model achieved comparable results to existing benchmarks while showcasing the advantages of lightweight and sequential fusion strategies.

In summary, our contributions are twofold:

1. We propose a Cascaded LoRA-based Fusion Framework for sequential and modular modality integration.
2. We validate the proposed architecture on two datasets, along with a comparative analysis against state-of-the-art methods to demonstrate proposed method performance.

The proposed framework establishes a foundational approach for dynamic sequential multi-modal learning. Integrating LoRA into this framework will allow for efficient and straightforward adaptation, minimizing the computational cost associated with introducing additional modalities while maintaining robust performance across tasks.

## 2 Related Work:

The growing interest in multimodal models stems from their capacity to capture a holistic understanding of complex environments by leveraging multiple data types or modalities. Such models enable a more nuanced interpretation and have been widely applied across diverse tasks. Notably, considerable research has been dedicated to the development of effective multimodal fusion techniques [11] [6].

Leveraging the advantages of cascaded network architectures, cascaded fusion techniques have been implemented across both transformer-based models [3] [8], as well as in non-transformer models,[2]. In parallel, the popular fine-tuning approach Low-Rank Adaptation (LoRA) [4] has emerged as a highly efficient method for multimodal tuning, especially due to its low computational demands and adaptability. LoRA-based tuning models are now widely used in various fusion architectures, encompassing late, early and mid-level including hierarchical fusion setups [13] [14] [9] [9] [12] [10].

Despite these advances, cascaded fusion models using LoRA have not yet been fully explored. We propose to develop a LoRA-based cascaded fusion model to leverage the flexibility of cascaded networks, which inherently support the incremental addition of new modalities. Apart from above, Our approach is close to hierarchical fusion method type, which share structural similarities with cascaded networks. Our approach even specially related to LoRA-based hierarchical multimodal techniques [15] [1] [16] [16].

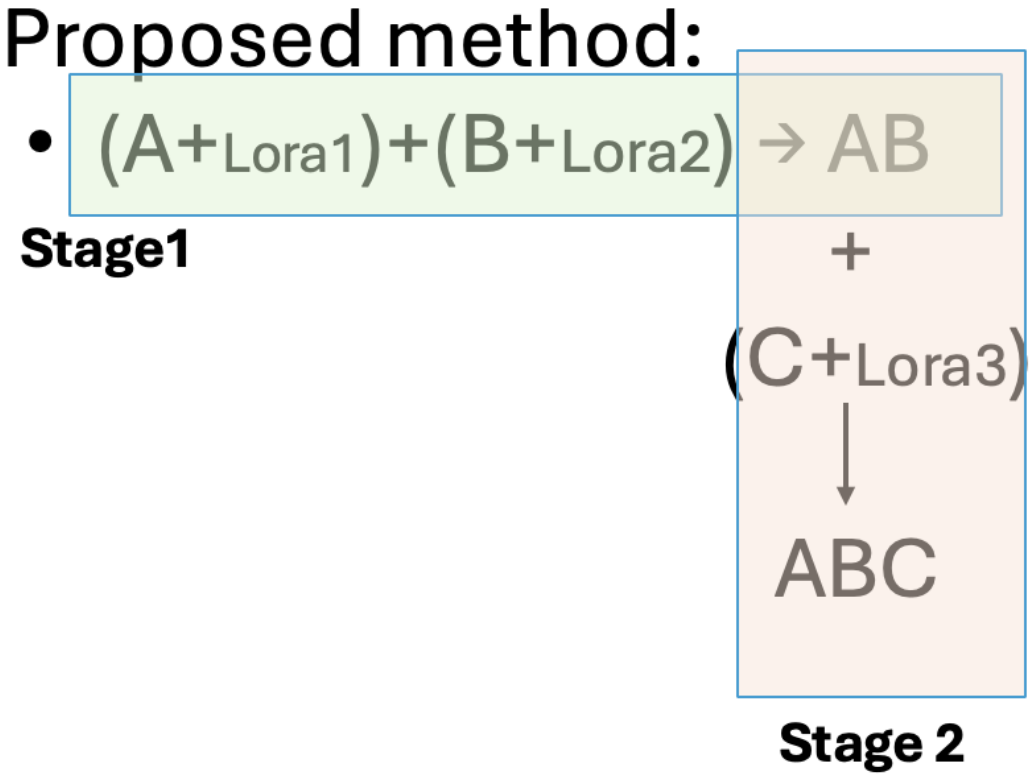

**Figure 1: Proposed Methodology-Cascaded Low-Rank Adaptation Multi-modal Fusion**

## 3 Proposed Methodology

### 3.1 Background

Low-Rank Adaptation (LoRA): LoRA is a fine-tuning method that focuses on updating low-rank matrices, offering an efficient and computationally lightweight alternative to traditional fine-tuning techniques. Instead of modifying the full model parameters, LoRA introduces low-rank updates to pretrained weights, significantly reducing memory and computational costs.

In the context of multimodality, LoRA brings distinct modality embeddings (e.g., text, video, and audio representations) closer together in the feature space, enabling effective modality-specific tuning. This capability is particularly advantageous for multimodal systems, where it allows efficient fine-tuning while maintaining the modularity of individual modalities.

Cascaded Multimodal Fusion: Cascaded fusion is a sequential integration technique that enables the step-by-step addition of modalities, such as $A + B \rightarrow AB \rightarrow AB + C \rightarrow ABC$ .The key advantage of this approach is that it avoids retraining the entire system when a new modality is added, thereby preserving the previously learned representations and reducing computational overhead.

This approach is particularly well-suited for evolving datasets where new modalities may be introduced over time. The cascaded fusion framework allows for flexibility in sequential modality addition, making it adaptable to dynamic multimodal learning scenarios.

### 3.2 Proposed Method

In this project, we aim to develop a LoRA-based cascaded fusion model that leverages the modularity and adaptability of cascaded networks to accommodate multiple modalities efficiently with the help of LORA. We conducted a detailed literature review to identify key multimodal fusion techniques, focusing on LoRA-based approaches and cascaded fusion architectures. While cascaded architectures available, their open source is code is not available, we are developing cascaded LORA architecture using three different singular modalities. The model architecture is based on cascaded fusion principles, where modality-specific LoRA adapters are sequentially integrated. Each modality (e.g., skeleton, RGB, text) is first processed through a modality-specific LoRA adapter for fine tuning purpose. Each LoRA adapter will be designed to process an individual modality. So overall all modalities are added sequentially (e.g., A+B $\rightarrow$ AB, then AB+C $\rightarrow$ ABC) without retraining prior components.

## 4 Experiments

### 4.1 Experimental Setup:

*4.1.1 Data:* We evaluate the proposed cascaded LoRA-based multimodal fusion framework on two healthcare-oriented action and activity recognition datasets: NurViD [5] and the Nurse Training dataset [7]. NurViD contains 1,538 untrimmed clinical videos totaling 144 hours, segmented into 5,608 clips with 177 action-step annotations. We focus on classification over these 177 action categories using synchronized video, audio, and MMPose-derived skeletal representations collected from real-world clinical environments.

The Nurse Training dataset[7] is a domain-specific nursing activity recognition benchmark containing 240 activity sequences from eight professional nurses performing six predefined activities. It includes motion-capture trajectories, Meditag-based indoor localization, and smartphone accelerometer data, capturing body movement, spatial location, air pressure, and motion dynamics. Due to its limited size and imbalanced activity durations, we use 3-fold cross-validation for evaluation.

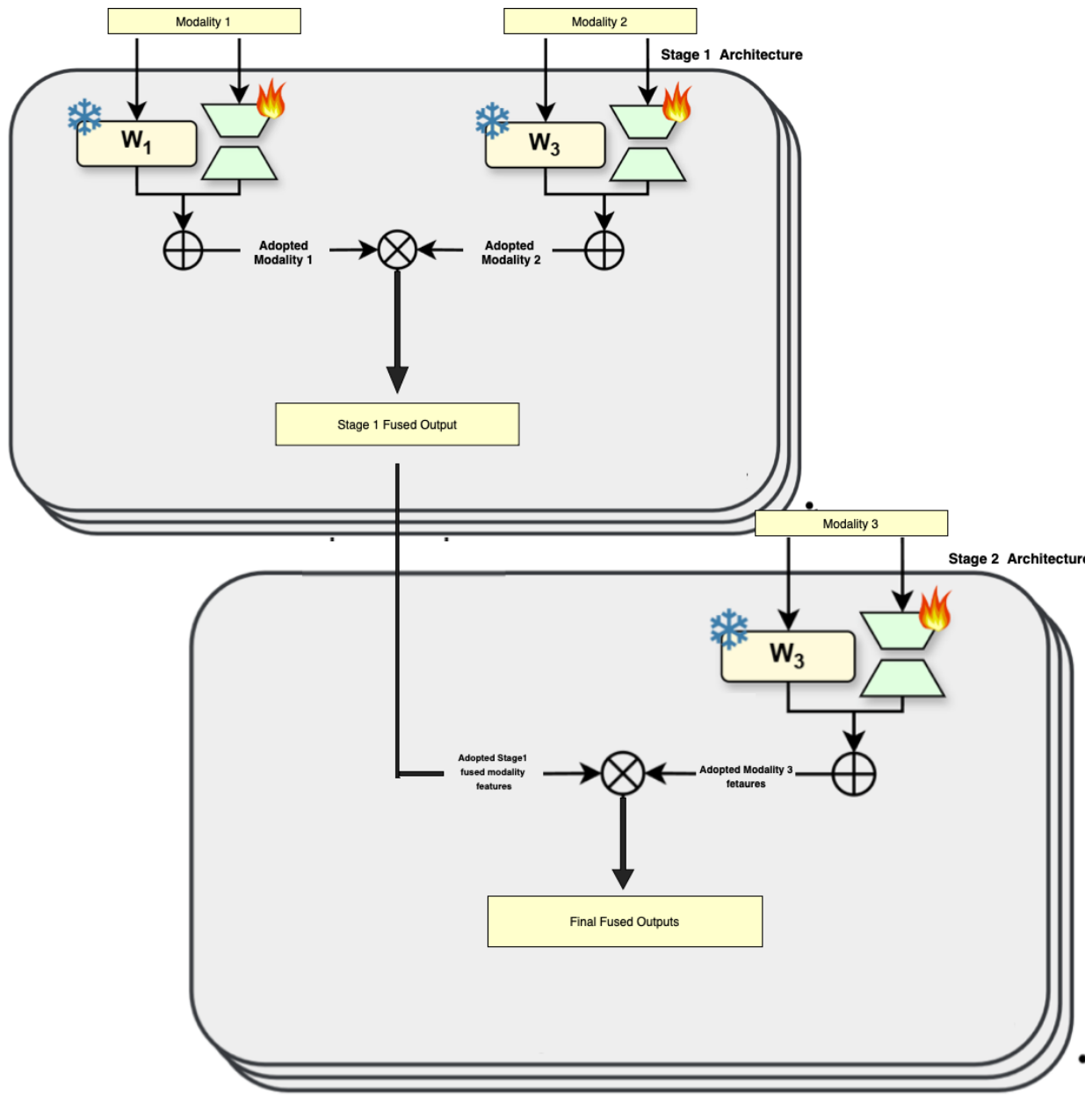

**Figure 2: Proposed Methodology-Cascaded Low-Rank Adaptation Multimodal Fusion**

## 4.2 Implementation Details

For Nurvid dataset, we developed a with a skeleton-only unimodel using STGCN finetuned model; RGB video features are modeled using TimeSformer; and audio is transcribed into text using google apIs and this modlaities processed using BERT.In case of Nurse training dataset, for position data, we developed a Meditag-based position transformer, inspired by the structure used for acceleration data. And [6] has already provided Acceleration Model(accelaration modalaity) and Spatio-Temporal Skeleton Model(skeleton sensors points modalaity). To validate our cascaded LoRA fusion model, we plan a series of experiments to evaluate its performance and flexibility. To validate our cascaded LoRA fusion model, we designed experiments tailored to the unique characteristics of two datasets: NurViD and Nurse Training Challenge. Each dataset utilizes distinct modalities and architectures, emphasizing the adaptability and performance of our proposed approach.

## NurViD Dataset:

The NurViD dataset, which focuses on action localization in untrimmed videos, incorporates the following three modalities: Skeleton Data:Skeleton data is processed using the Nurse Training Transformer (skeleton-only variant) fine-tuned with LoRA layers, capturing spatial-temporal patterns critical for action localization. Video Frames (RGB)" Video frames are modeled using TimeSformer, which effectively extracts spatial-temporal features from video data, aiding in precise motion representation. Text Data: Audio data is transcribed into text using the Google API, and the resulting text is encoded with BERT, fine-tuned with LoRA layers to extract semantic information relevant to activity localization.

## Nurse Training Challenge Dataset:

The Nurse Training Challenge dataset, focused on nursing activity recognition, utilizes the following three modalities: Skeleton Data: Skeleton data is modeled using the Skeleton Transformer from the Nurse Training Transformer paper, capturing movement patterns associated with nursing tasks. Acceleration Data: Acceleration data, collected via smartphones placed in nurses' chest pockets, is processed using a basic transformer architecture, fine-tuned with LoRA layers to detect activity-specific motion.Position Data: Positional data, obtained via Meditag-based indoor localization, is processed using a Meditag-based position transformer, designed to extract spatial insights similar to the acceleration transformer.

*4.2.1 Evaluation Plan:* We utilize accuracy as the primary metric to evaluate the effectiveness of our model in classifying activities based on sequentially fused modalities.

*4.2.2 Baseline Comparisons:* We will establish a baseline by testing the unimodal version of the model and comparing it to fully fine-tuned multimodal models that integrate all available modalities without LoRA. This provides a reference for the benefits of cascaded LoRA fusion.

## 4.3 Experiments and Results:

*4.3.1 Nurse Training Dataset.* Table 1 reports the performance of individual modalities, late fusion, and the proposed two-stage cascaded fusion strategy on the Nurse Training dataset. Among the unimodal models, skeleton features achieve the strongest performance with 0.71 accuracy, while acceleration and Meditag position features obtain 0.38 and 0.30 accuracy, respectively. Although acceleration and position are weaker as standalone modalities, their contribution becomes evident when integrated with skeleton features. Stage 1 fusion, which combines skeleton and acceleration modalities, improves accuracy to 0.7586, matching the reported cross-modal fusion baseline. Stage 2 further incorporates Meditag position features and improves performance to 0.7931, outperforming both unimodal and late-fusion configurations. These results suggest that the proposed LoRA-based cascaded fusion framework effectively integrates complementary sensor modalities while maintaining a lightweight and sequential fusion structure.

**Table 1: Performance Metrics for Different Model Stages Nurse Care Activity Recognition Challenge Dataset**

| Model | Accuracy (%) |
|---|---|
| Simple Fusion(Late Fusion of Three Modalities) | 0.73 |
| Skeleton Sensors Data Only (Modlaity1) | 0.71 |
| Acceleration Sensor Data Only (Modlaity2) | 0.38 |
| Position Sensor Data Only - Meditag (Modality3) | 0.3 |
| Stage1 Fusion (Modality1+Modlaity2) | 0.7586 |
| Stage2 Fusion (Stage1+Modlaity3) | 0.7931 |

*Note.* all the models are finetuned on Nurse care activity recognition challenge dataset

#### 4.3.2 *NurViD Dataset.*

For the NurViD dataset, we evaluate the proposed framework using individual modality models and two sequential fusion stages. As shown in Table 2, RGB achieves the strongest unimodal performance, followed by text and skeleton representations. In the first stage, we fuse the skeleton and RGB modalities, following multimodal learning literature that suggests integrating more closely related modalities before incorporating more heterogeneous modalities. This design choice is also motivated by the fact that skeleton representations are derived from RGB frames and therefore share stronger spatial-temporal correspondence with the visual modality than with audio-derived text. In the second stage, we incorporate the text modality obtained from transcribed audio. Accuracy improves from 40.46% for the best unimodal model to 61.87% after Stage 1 fusion and 83.02% after all-modality fusion, indicating that the proposed parameter-efficient cascaded fusion strategy provides a promising direction for sequentially fusing multiple modalities in the context of action recognition for medical training environments.

**Table 2: Performance Metrics for Different Model Stages**

| Model | Test Accuracy (%) |
|---|---|
| Skeleton Model | 11.25 |
| RGB Model | 37.37 |
| Text Model | 38.11 |
| Stage 1 Fusion( skeleton & Vision) | 61.70 |
| Stage 2 Fusion (All Modalities) | 83.02 |

*Note.* all the models are finetuned on nurvid data

## 5 State-of-the-Art Comparison

**Table 3: Comparison with state-of-the-art methods on Nurse Training Dataset**

| Model | Accuracy(%) |
|---|---|
| Cross Fusion Multi-model Transformer (MNT [7]) | 75.8 |
| LoRA-Based Cascaded Multimodal Fusion (Ours) | 79.31 |

**Table 4: Comparison with state-of-the-art methods on Nurvid Dataset**

| Model | Accuracy(%) |
|---|---|
| SlowFast* (baseline from [5]) | 14.83 |
| C3D* (baseline from [5]) | 14.90 |
| I3D* (baseline from [5]) | 14.83 |
| LoRA-Based Cascaded Multimodal Fusion (Ours) | 83.02 |

*Note.* An asterisk (*) denotes the initialization from the model pre-trained on Kinetics 400

For the Nurse Training dataset, the prior MNT model [7] reports 75.8% accuracy using skeleton and acceleration modalities. Our Stage 1 cascaded fusion model uses the same modalities and achieves comparable performance at 75.86%. As shown in Table 3 by further incorporating Meditag-based position data in Stage 2, the proposed model improves accuracy to 79.31%, demonstrating the benefit of sequentially integrating complementary sensor modalities.

For NurViD as shown in Table 4 , we compare against the baselines reported in [5]. SlowFast, C3D, and I3D achieve 14.83%, 14.90%, and 14.83% accuracy, respectively, whereas our LoRA-based cascaded multimodal fusion model achieves 83.02% by integrating skeleton, RGB, and audio-derived text modalities. Overall, these results show that the proposed framework provides competitive or improved performance while supporting parameter-efficient and sequentially scalable multi-modal fusion.

## 6 Conclusion

In this paper, we presented a cascaded Low-Rank Adaptation (LoRA)-based multimodal fusion framework for action and activity recognition in healthcare-oriented training environments. The proposed framework combines cascaded fusion with parameter-efficient LoRA fine-tuning, enabling modalities to be integrated sequentially without retraining previously learned components. This design supports modularity, computational efficiency, and adaptation across datasets with different modality configurations.

We evaluated the framework on NurViD and the Nurse Training dataset. The results suggest that sequentially integrating complementary modalities can improve performance over individual modality models and provide competitive performance relative to previously reported dataset-specific baselines. In particular, the framework demonstrates the value of first integrating closely related modalities and then incorporating more heterogeneous modalities in later fusion stages.

Following LoRA, MixLoRA, and LoRA-based fusion literature, modality-specific adapters encourage representations to become more closer in the latent space, reducing the need for costly cross-fusion mechanisms and enabling simple fusion operations such as concatenation. Future ablation studies will further examine the effects of LoRA adaptation, fusion order, resource-utilization and runtime optimizations. Overall, the proposed cascaded LoRA-based fusion framework offers a promising direction for scalable and parameter-efficient multimodal fusion in medical training environments.

## 7 Limitations and Applicability

The proposed cascaded LoRA-based multimodal fusion framework efficiently aggregates complementary information from different modalities; however, it has several limitations. First, it relies on implicit co-adaptation of modality representations by fusing output embeddings from modality-specific architectures. Thus, the approach may be most suitable when moderate representation refinement is sufficient; settings requiring stronger modality alignment need further ablation studies. Second, the sequential fusion strategy may introduce a representation bottleneck, since later modalities must adapt to partially fixed representations from earlier stages. While this improves modularity and efficiency, it may limit globally optimal multimodal representation learning when strong cross-modal dependencies exist.Despite these limitations, the framework is well suited for classification-oriented tasks with complementary modalities, evolving multimodal inputs, and computational constraints. Future work should examine alignment-aware mechanisms for tasks requiring explicit cross-modal interaction modeling.